\documentclass{article}

     \usepackage[preprint]{neurips_2021}

\usepackage[utf8]{inputenc} 
\usepackage[T1]{fontenc}    
\usepackage{hyperref}       
\usepackage{url}            
\usepackage{booktabs}       
\usepackage{amsfonts}       
\usepackage{nicefrac}       
\usepackage{microtype}      
\usepackage{xcolor}         

\usepackage{amsmath, amsthm, amssymb, bm} 
\usepackage{algorithm}
\usepackage{algorithmic}
\usepackage{graphicx} 
\graphicspath{{figures/}} 
\usepackage[font=small,labelfont=bf]{caption} 

\usepackage[utf8]{inputenc}
\usepackage[T1]{fontenc}

\title{2nd-order Updates with 1st-order Complexity}

\author{%
  Michael F.~Zimmer \\
  \texttt{zim@neomath.com} \\
}

\begin{document}

\maketitle

\begin{abstract}
It has long been a goal to efficiently compute and use second order information on a function ($f$) to assist in numerical approximations.
Here it is shown how, using only basic physics and a numerical approximation, such information can be accurately obtained
at a cost of ${\cal O}(N)$ complexity, where $N$ is the dimensionality of the parameter space of $f$.
In this paper, an algorithm ({\em VA-Flow}) is developed to exploit this second order information, and pseudocode is presented.
It is applied to two classes of problems, that of inverse kinematics (IK) and gradient descent (GD).
In the IK application, the algorithm is fast and robust, and is shown to lead to smooth behavior even near singularities.
For GD the algorithm also works very well, provided the cost function is locally well-described by a polynomial.
\end{abstract}

\section{Introduction}

The ability to compute second order information of a function adds obvious benefits over only using first order information \citep{Bishop-book2006}.
Two obvious applications are in optimization (gradient descent (GD)) and in robotics (inverse kinematics (IK)).
To date, the main challenge has been the complexity costs (i.e., computations and storage), which are ${\cal O}(N^2)$
when there are $N$ parameters; in contrast, first-order computations have ${\cal O}(N)$ complexity.
This paper will introduce an algorithm which extracts the benefits of a second order algorithm,
but only incurs ${\cal O}(N)$ complexity.

After discussing relevant research in the section {\em Related Work},
the new ideas and algorithms are introduced in the sections {\em Vector Flows}, {\em Usage},
and {\em The VA-Flow Algorithm}.
The following three sections contain a simple demonstration on an ellipse,
numerical experiments on IK and GD, as well as a concluding discussion.
Finally, the reader will find additional material in the Supplement and on the author's Github site \citep{MFZ-github}.

\section{Related Work}
\label{sec:background}

Work that relates to second order corrections (e.g., Hessians) and to applications that make use of such corrections
(e.g., GD and IK) are of primary interest here.  In the {\em Summary} sub-section, a general characterization of the applications
is given, which will be used throughout this paper.

\subsection*{Inverse Kinematics (IK)}

This application concerns itself with adjusting the pose of an articulated figure \citep[e.g., see][]{Sciavicco2001,Buss2009,Aristidou2009}.
For concreteness, if a human wishes to move his hand from point A to point B, while keeping his torso fixed,
the problem becomes one of determining the changes in the orientation of the shoulder, elbow, and wrist.
More abstractly, the arm is described as {\em links} (the bones of the arm) connected by {\em joints} (e.g., the elbow),
and the standard IK problem is to determine how to adjust the joints in order to move the end of the last link (the {\em end-effector}) from point A to B.
The position of end-effector may be formulated as
\begin{align*}
{\bm r} = f( {\bm \theta} ) \, \, \, ,
\end{align*}
where ${\bm \theta} = (\theta_1, \dots, \theta_N)$ is a parameterization of the $N$ joints, and ${\bm s}$ is the position of the end-effector.

Following \cite{Aristidou2009}, there are four main classes of solutions for this problem:
Jacobian inverse, Newton, Heuristics, and Data-driven.
(Heuristic and Data-Driven methods are not of direct interest for this paper, and will not be discussed.)
The Jacobian inverse method centers on making small changes $d{\bm r}$, $d{\bm \theta}$, and are related by
\begin{align}
d{\bm r} = {\bm J} d{\bm \theta} \, \, \, ,
\label{eqn:IKdr}
\end{align}
where ${\bm J} = df/d{\bm \theta}$ is the Jacobian.
(For example, if the goal is to move the end-effector from ${\bm r}_A$ to ${\bm r}_B$, one could define
${\bm s} = {\bm r}_B - {\bm r}_A$ and $\hat{ {\bm s} } = {\bm s} / \| {\bm s} \|$
so that $d{\bm r} = \alpha \hat{ {\bm s} }$.
Here $\alpha$ is the magnitude of the move, which may be taken as the minimum of $\| {\bm s} \|$
and a user-defined maximum.)
Eq.~\ref{eqn:IKdr} is normally an underdetermined system of equations; it additionally offers the challenge of 
singular points in ${\bm \theta}$-space, which are associated with a change in rank of the matrix $J$.
Some of the approaches that have appeared in this class of solutions are:
Jacobian pseudoinverse \citep{Whitney1969},
damped least-squares \citep{Wampler1986,Nakamura1986}, 
singular value filtering \citep{Colome2012},
pseudoinverse damped least-squares \citep{Maciejewski1990,Press-book2007},
selectively damped least-squares \citep{Buss2005}, and others.
Of these named methods, the fastest (and least robust) is the Jacobian pseudoinverse, and will be the baseline for comparisons herein.  
It obtains a solution in the form \citep{MFZ-AXB}
\begin{align*}
d{\bm \theta} & = {\bm G} d{\bm r} + d{\bm \theta}_{null} \\
{\bm P} & = 1 - {\bm G} {\bm J} \, \, \, ,
\end{align*}
where $d{\bm \theta}_{null}$ is in the null space of ${\bm J}$, ${\bm G}$ is the generalized inverse \citep{BenIsrael-book} of ${\bm J}$, and ${\bm P}$ is a projection operator for the null space of ${\bm J}$.
When ${\bm G}$ is of type \{1234\}, it is known as a Moore-Penrose inverse (often written as ${\bm J}^\dagger$).
The operator ${\bm P}$ can be used for secondary goals \citep{Liegeois1977} of the articulated arm, such as satisfying joint constraints.
Also, Newton methods give an accurate 2nd-order method, but are computationally very expensive to implement.
In summary, the challenge with the IK problem is that existing algorithms have been either fast or robust, but not both.

IK algorithms find applications wherever articulated limbs are used, such as computer graphics and robotic arms (in manufacturing, human-like robots, arms in space travel, prosthetic limbs, etc.).  Thus any opinion of the ethics of this application is simply a reflection of the ethics of those existing technologies, which in the author's opinion has been positive thus far.

\subsection*{Gradient Descent (GD)}

Gradient descent \citep{Cauchy1847,Lemarechal2010,Press-book2007} is an iterative optimization algorithm to be used for the minimization of a differentiable function.  
With respect to the cost function (CF) $f$, which depends on the parameter vector
${\bm \theta} = (\theta_1, \theta_2, ...)$, an initial value ${\bm \theta}_{old}$ is updated to ${\bm \theta}_{new}$ as
\begin{align*}
{\bm \theta}_{new} & = {\bm \theta}_{old} - \alpha {\bm \nabla} f \, \, \, ,
\end{align*}
where ${\bm \nabla} f$ is the gradient of $f$ with respect to ${\bm \theta}$, and $\alpha$ is a positive number known as the learning rate.

The topic of iterative approaches to optimization has been well investigated
(e.g., see \citep{Nesterov-book2018,Sra-book2011,Nocedal-book2006}),
and most methods are classified as either line search or trust region approaches.
Other useful techniques include momentum and stochastic gradient descent \citep[e.g., see][]{Goodfellow-book2016}.

There is a closely related class of algorithms that involve momentum \citep{Polyak1964,Nesterov1983} being applied to the gradient and/or gradient-squared,
effectively creating a different learning rate for each component of ${\bm \theta}$.
In this category are the algorithms AdaGrad \citep{Duchi2011}, AdaDelta \citep{Zeiler2012}, RMS Prop \citep{Tieleman2012}, Adam \& AdaMax \citep{Kingma2015},
Nadam \citep{Dozat2016}, AdamNC \citep{Reddi2018},
and others \citep[see][] {Ruder2017}.
At this point, Adam seems to be the preferred choice, and it will be used for comparisons herein.
These algorithms were originally meant to be used with stochastic gradient descent, but may be used in batch GD as well.

Additionally, the recent Neograd algorithm \citep{MFZ-neograd} formulaically adjusts the learning rate to keep a metric near its target value.  
It uses the philosophy that the deviation from the first-order update should be relatively small compared to the first-order update itself. 
In batch optimizations compared to Adam, it was shown to reach lower CF values by factors such as $10^8$ to $10^{12}$.

Finally, GD finds use in myriad applications, one of them being supervised learning for image detection.
That application has ethically positive uses (e.g., detecting tumors from X-ray images), as well as
dystopian ones (e.g., pervasive facial recognition technology in society).
Thus one will have to survey all such applications to decide whether the net ethical impact
is good or bad.

\subsection*{Hessians}

In a Taylor series expansion of a function $f$, the Hessian ($H$) appears as the matrix in the second order term.
If $f$ depends on $N$ parameters, then $H$ has $N^2$ terms, and thus a direct computation of $H$ becomes impracticable for large N.
A quasi-Newton method \citep[cf. ][]{Nocedal-book2006} can be used as an approximation toward computing it,
and then determining its impact.
Other approximations for computing $H$ include Broyden’s method, Powell’s method,
Siciliano’s method, as well as the BFGS method and its variants \citep[see][]{Fletcher1987, Siciliano1990, Chin1997}.
In general, these methods are known for being accurate and more stable, but computationally costly.

In two independent efforts, \cite{Moller1993,Moller1993-thesis} and \cite{Pearlmutter1994} determined a way to use $H$ without explicitly calculating and storing its $N^2$ entries.
Noting that the Hessian normally appears as $H {\bm w}$, where ${\bm w}$ is a suitable vector, 
they each observed that it was accessible by a numerical approximation via
$H {\bm w} \approx (g( {\bm \theta} + \epsilon {\bm w}) - g( {\bm \theta} ))/\epsilon$, where $g( {\bm \theta} )$ is the gradient of $f$ and $\epsilon$ is a small number.
While they primarily employed this in the context of neural nets,
they both also discussed how it can be used in the line search algorithm to set the step size ($\alpha$), such as with scaled conjugate gradient \citep{Moller1993-SCG, Moller1993-SSCG}.
They both also noted its use in estimating eigenvalues and eigenvectors of $H$.

\subsection*{Summary}

The applications of IK and GD may be treated side by side by identifying the vector field ${\bm v}({\bm \theta})$ 
\begin{equation}
\renewcommand{\arraystretch}{1.2}
{\bm v} ( {\bm \theta} ) = \left\{
\begin{array}{@{\quad}l@{\quad}l@{}}
-{\bm \nabla} f |_{\bm \theta}  &  \text{ (gradient descent) } \\
G( {\bm \theta}) \hat{ {\bm s} }  &  \text{ (inverse kinematics) } , \\
\end{array}
\right.
\label{eqn:v-summary}
\end{equation}
where all quantities are as defined above.
In the algorithms of interest the parameter ${\bm \theta}$ is iteratively updated by the amount
\begin{align*}
d{\bm \theta} = \alpha {\bm v} \, \, \, ,
\end{align*}
where $\alpha$ is a positive number.

\section{Vector Flows}
\label{sec:curves}

Computing updates of ${\bm \theta}$ with a single vector ${\bm v}$ is convenient and quick,
but a more accurate prediction would involve the rate of change of ${\bm v}$.
Here a numerical estimate is made of the rate of change of ${\bm v}$ at the two nearby values ${\bm \theta}_0$ and ${\bm \theta}_1$.
Instead of making a single update using $\alpha$, a smaller update using $\epsilon$ is made.  
In this paper this value is set as 
\begin{align*}
\epsilon & = \alpha/M \ll \alpha \, \, \, ,
\end{align*}
where $M$ is a suitably large number (cf. Sec.~\ref{sec:va-flow}).
This allows a calculation of what is called the {\em acceleration}
\begin{align}
{\bm a} & \equiv ( {\bm v}_1 - {\bm v}_0 ) / \epsilon \, \, \, ,
\label{eqn:acc}
\end{align}
where ${\bm v}_k = {\bm v} ({\bm \theta}_k)$ and ${\bm \theta}_1 = {\bm \theta}_0 + \epsilon {\bm v}_0$.
Next, the compounded effect of ${\bm a}$ on the scale of $\epsilon$ is computed, to determine its influence on the scale of $\alpha$.
Assuming that ${\bm a}$ remains constant for these subsequent updates, the net effect can be computed as (with $n \geq 1$)
\begin{subequations}
\label{eqn:n-updates}
\begin{align}
{\bm v}_n & = {\bm v}_{n-1} + \epsilon {\bm a}  \nonumber \\
& = {\bm v}_0 + n\epsilon {\bm a} \label{eqn:n-updates-v} \\
{\bm \theta}_n & = {\bm \theta}_{n-1} + \epsilon {\bm v}_{n-1} \nonumber \\
 & = {\bm \theta}_0 + \epsilon ({\bm v}_0 + \cdots + {\bm v}_{n-2} + {\bm v}_{n-1} ) \nonumber \\
 & = {\bm \theta}_0 + n \epsilon {\bm v}_0 + \frac{1}{2} n(n-1) \epsilon^2 {\bm a} \label{eqn:n-updates-theta} \, \, \, .
\end{align}
\end{subequations}
Assuming large $n$, and identifying $\alpha = n \epsilon$, it approximately follows that 
\begin{subequations}
\label{eqn:updates}
\begin{align}
{\bm v}_\alpha & = {\bm v}_0 + \alpha {\bm a}  \label{eqn:update1} \\
{\bm \theta}_\alpha & = {\bm \theta}_0 + \alpha {\bm v}_0 + \frac{\alpha^2}{2} {\bm a} \label{eqn:update2} \, \, \, .
\end{align}
\end{subequations}
These two {\em update formulas} should be immediately recognizable as two equations from elementary physics:
the former for the velocity ${\bm v}$ due to an acceleration ${\bm a}$ over a time $\alpha$ given an initial velocity ${\bm v}_0$,
and the latter for the change in position ${\bm \theta}$ due to the same.

\subsection{Further Aspects}

This section collects several incidental results related to the previous section.
Eq.~\ref{eqn:acc} can be written as 
\begin{subequations}
\label{eqn:acc-taylor}
\begin{align}
{\bm a} 
& = [ {\bm v} ({\bm \theta}_0 + d\hat{ {\bm \theta} })  - {\bm v} ({\bm \theta}_0) ] / \epsilon \nonumber \\
& \approx ( d\hat{ {\bm \theta} } \cdot {\bm \nabla} ) {\bm v} / \epsilon \nonumber \\
& \approx ( {\bm v}_0 \cdot {\bm \nabla} ) {\bm v}  \label{eqn:acc-taylor-3} \, \, \, ,
\end{align}
\end{subequations}
where $d\hat{ {\bm \theta} } = \epsilon {\bm v}_0$.
Evaluations of the derivative are understood to be made at ${\bm \theta}_0$.
Note that in the context of GD where ${\bm v} = - {\bm \nabla} f$, ${\bm a}$ becomes $ {\bm \nabla} f ( {\bm \nabla} {\bm \nabla} f)$,
which was studied by \cite{Moller1993,Moller1993-thesis} and \cite{Pearlmutter1994}.
Using this expression for ${\bm a}$, and now setting $d {\bm \theta} = \alpha {\bm v}_0 + \alpha^2 {\bm a}/2$,
a second-order approximation for the CF $f( {\bm \theta} )$ can be written as
\begin{subequations}
\label{eqn:ftaylor-all}
\begin{align}
f & = f_0 + ({\bm \nabla} f)^T d{\bm \theta} + \frac{1}{2} d{\bm \theta}^T ( {\bm \nabla}{\bm \nabla}f  ) d{\bm \theta} + {\cal O}( \| d{\bm \theta} \|^3 ) \nonumber \\
 & = f_0 - \alpha v_0^2  - \alpha^2 {\bm v}_0 \cdot {\bm a}  + {\cal O} (\alpha^3) \, \, \, , \label{eqn:ftaylor}
\end{align}
\end{subequations}
where $v_0^2 = {\bm v}_0 \cdot {\bm v}_0$, and $f_0 = f({\bm \theta}_0)$.  
Note the improvement in estimating $f$ due to the term $- \alpha^2 {\bm v}_0 \cdot {\bm a}$.

Finally, a differential geometric \citep[e.g., see][]{Lee-book2012,Frankel-book2019,Tu-book2006} interpretation of ${\bm a}$ may be made, beginning from an expansion of the vector field ${\bm v}$ at ${\bm \theta}_0$:
\begin{align}
{\bm v} = {\bm v}_0  + ( d {\bm \theta} \cdot {\bm \nabla} ) {\bm v}  + \cdots \, \, \, .
\label{eqn:v-taylor}
\end{align}
Here $d {\bm \theta} = {\bm \theta} - {\bm \theta}_0$ is the only ${\bm \theta}$ dependence on the right hand side, since the gradient is evaluated at ${\bm \theta}_0$.
In the neighborhood of ${\bm \theta}_0$, define the vector fields $X$ and $Y$ based on the contributions from the first and second order terms of Eq.~\ref{eqn:v-taylor}
\begin{align*}
X & = {\bm v}_0 \cdot {\bm \nabla} \\
Y & = {\bm v}_0 \cdot {\bm \nabla} + [ (d {\bm \theta} \cdot {\bm \nabla} ) {\bm v} ] \cdot {\bm \nabla} \, \, \, .
\end{align*}
The vector field $X$ represents basic GD, while $Y$ also includes the contribution from a changing ${\bm v}$.
Thus we are interested in the rate of change of $Y$ due to the flow created by $X$.
This naturally motivates the consideration of the Lie derivative \citep[introduced by][]{Slebodzinski1931} of $Y$ with respect to $X$,
which when evaluated at ${\bm \theta}_0$ is
\begin{align*}
{\cal L}_X Y & = [X, Y] \\
& = [ ( {\bm v}_0 \cdot {\bm \nabla} ) {\bm v} ] \cdot {\bm \nabla} \\
& = {\bm a} \cdot {\bm \nabla} \, \, \, ,
\end{align*}
where the last line made use of Eq.~\ref{eqn:acc-taylor-3}.
Thus, the vector ${\bm a}$ in the small-$\epsilon$ limit of Eq.~\ref{eqn:acc} may also be understood to be equivalent to the Lie derivative ${\cal L}_X Y$.
This entire procedure could be repeated for higher order corrections in Eq.~\ref{eqn:v-taylor}.
Also, note that the original expression for ${\bm a}$ (i.e., Eq.~\ref{eqn:acc}) was valid because the underlying space is Euclidean.
Thus even though ${\bm v}_0 $ and ${\bm v}$ live in different tangent spaces and in general can't be directly compared,
they can be in this case (cf. p.228 in \cite{Lee-book2012}).

\section{Usage}

This section introduces several heuristics for using the second order update formulas in Eqs.~\ref{eqn:n-updates}.
They're based on the principal that the correction to the first-order update should be smaller than the first-order update itself \citep[cf.][]{MFZ-neograd}.
Note that this is different from line search techniques, which attempt to lower the CF as much as possible in a single update.
(Although, Wolfe conditions \citep{Armijo1966,Wolfe1969,Wolfe1971} are a refinement on this.)

\subsection{Approach A}

The metric $\rho_A$ is defined as the ratio of the magnitudes of the second term to the first from Eq.~\ref{eqn:n-updates-theta}
\begin{align*}
\rho_A & =  \frac{  \frac{1}{2} n(n-1) \epsilon^2  \| {\bm a} \| }{ n\epsilon \| {\bm v} \| }
 = \frac{ (n-1) \epsilon }{2} \frac{ \| {\bm a} \| }{ \| {\bm v} \| } \, \, \, .
\end{align*}
This equation provides a way to solve for the value of $n = n^*$ that leads to a specified value
for $\rho_A$, which in this case is the target value $\rho_{targ}$
\begin{align}
n^* & = 1 + \frac{ 2\rho_{targ} }{\epsilon} \frac{ \| {\bm v} \| }{ \| {\bm a} \| } \, \, \, .
\label{eqn:nstarA}
\end{align}
This value of $n^*$ reveals how many 
\footnote{The value of $n^*$ should be cast as an integer, but that restriction isn't necessary for its use here.}
updates should be taken to maintain the metric at the value $\rho_{targ}$.
This value should be substituted into Eq.~\ref{eqn:n-updates-theta} to obtain the new parameter value.
Thus, this approach provides for a {\em single update} $\theta_{n^*}$, starting from $\theta_0$.
Note that for this approach to make sense, $n^* \gg 1$; if not then then it's essentially just GD.
This is the preferred approach in this paper.

\subsection{Approach B}

The approach here is similar to that for $\rho_A$, except now the contributions from the right-hand side of Eq.~\ref{eqn:n-updates-theta}
are measured along the direction of $\hat{ {\bm v} } = {\bm v} / \| {\bm v} \|$.  The metric is
\begin{align*}
\rho_B & = \left| \frac{ \hat{{\bm v}} \cdot [ \frac{ n(n-1)\epsilon^2 }{2} {\bm a} ] }{ \hat{{\bm v}} \cdot [ n\epsilon {\bm v} ] } \right|
= \frac{ (n-1)\epsilon }{2} \frac{ | {\bm v} \cdot {\bm a} | }{ v^2 } \, \, \, .
\end{align*}
Note this is the same (up to a factor of 2) as taking the ratio of the second to the first correction in Eq.~\ref{eqn:ftaylor}.
As before, this equation provides for a way to solve for the value of $n = n^*$ that leads to the value $\rho_B = \rho_{targ}$
\begin{align*}
n^* & = 1 + \frac{ 2\rho_{targ} }{\epsilon} \frac{v^2}{ | {\bm v} \cdot {\bm a} | } \, \, \, .
\end{align*}
This value should be used in the same manner as with {\em Approach A}.
Finally, in the context of GD where ${\bm a} = ( {\bm v}_0 \cdot {\bm \nabla} ) {\bm v}$ and ${\bm v} = -{\bm \nabla} f$, it easily follows that in the small $d{\bm \theta}$ limit, it becomes identical to the metric from \cite{MFZ-neograd}
\begin{align*}
\rho_B 
 & = \frac{1}{2} \left| \frac{ d{\bm \theta}^T ({\bm \nabla} {\bm \nabla} f ) d{\bm \theta} }{ {\bm \nabla} f^T d{\bm \theta} } \right| \, \, \, .
\end{align*}
That metric was also based on keeping the deviation from the first-order correction small.

\subsection{Approach C}

In a similar manner, comparing the second to the first terms of the right hand side of Eq.~\ref{eqn:n-updates-v} leads to the metric
\begin{align*}
\rho_C = \frac{ \| n \epsilon {\bm a} \| }{ \| {\bm v} \| } \, \, \, .
\end{align*} 
In this case the number of updates of size $\epsilon$ needed to reach $\rho_C = \rho_{targ}$ is
\begin{align*}
n^* = \frac{\rho_{targ} }{ \epsilon } \frac{ \| {\bm v} \| }{ \| {\bm a} \| } \, \, \, .
\end{align*} 
For large $n^*$, this becomes equivalent to Eq.~\ref{eqn:nstarA}, except it is missing the factor of 2.

\section{The VA-Flow Algorithm}
\label{sec:va-flow}

Pseudocode for the ideas discussed thus far is given below in Algo.~\ref{algo:va-flow} in what is called 
the {\em VA-Flow} algorithm.
Note that lines 4-7 correspond to the computations of ${\bm v}$ and ${\bm a}$ in Sec.~\ref{sec:curves}, with
the function {\em get\_v} implementing the appropriate choice from Eq.~\ref{eqn:v-summary}.
Line 9, which includes the function {\em get\_dtheta}, implements Eq.~\ref{eqn:n-updates-theta};
{\em get\_nstar} implements Eq.~\ref{eqn:nstarA}.  (Approach B or C could also be used).
These are the most essential elements of the algorithm.
The parameter values generally  used by the author were $\rho_{targ} = 0.1$ and $M=100$; the value of $num$ depends on the application.
(This value of $M$ was chosen to provide room for $n^*$ to adjust while still keeping $n^* \gg 1$.)
\begin{algorithm}[H]
\caption{: VA-Flow, version 0}
\begin{algorithmic}[1]
\STATE{Input: (${\bm \theta}$, $\alpha$, $\rho_{targ}$, $M$, $num$ ) }
\FOR{i = 1 to num}
\STATE{ $\epsilon = \alpha / M$ }
\STATE{ ${\bm v} = \text{get\_v} ( {\bm \theta} )$ }
\STATE{ ${\bm \theta}_1 = {\bm \theta} + \epsilon {\bm v}$ }
\STATE{ ${\bm v}_1 = \text{get\_v} ( {\bm \theta}_1 )$ }
\STATE{ ${\bm a} = ( {\bm v}_1 - {\bm v} ) / \epsilon$ }
\STATE{ $n^* = \text{get\_nstar} ( {\bm v}, {\bm a}, \rho_{targ}, \epsilon ) $ }
\STATE{ ${\bm \theta} = {\bm \theta} + \text{get\_dtheta} ( {\bm \theta}, {\bm v}, {\bm a}, n^*, \epsilon ) $ }
\STATE{ $\alpha = n^* \epsilon$ }
\ENDFOR
\STATE{Return: ${\bm \theta}$ }
\end{algorithmic}
\label{algo:va-flow}
\end{algorithm}
The reader should note the interplay between the statements on lines 3 and 10, which allow $\alpha$ and $\epsilon$ to adjust.
On line 10, $\alpha$ is reset using the most recent $n^*$ value,
and on line line 3, $\epsilon$ is again rescaled to be much smaller than $\alpha$.
It is important that $n^* \gg 1$, to conform to the approach given in Sec.~\ref{sec:curves}.
If $n^* \approx 1$, then it is effectively the same as GD, and no second-order benefits can be expected.
Also, the reader may instead use get\_dtheta to simply make a first-order update of ${\bm \theta}$;
in this case, this algorithm is only used as a means to set $\alpha$ in an otherwise first-order GD.
Finally, notice that VA-Flow is an ${\cal O}(N)$ algorithm, since the additional operations beyond GD in lines 6-9 are all ${\cal O}(N)$.
(It is assumed that get\_v is ${\cal O}(N)$.)

Note that the benefit of using ${\bm a}$ is perhaps greatest when $\| {\bm a} \| / \| {\bm v} \| $ is relatively large.
It is indicative of a rapidly changing ${\bm v}$, and thus provides important information regarding the next update of ${\bm \theta}$.
In contrast, when that ratio is relatively small ${\bm a}$ becomes less important, and it can lead to large values of $n^*$.
In this case it may be necessary to use other information to limit the size of $\alpha$.
(Alternatively, one could simply default to the underlying first-order algorithm.)
It will be shown that for IK and GD on a polynomial CF it is not an issue, but it can be for GD on a CF with exponential terms.

Finally, there are modifications that can be made to improve performance and robustness, and they are discussed in the Supplement.
There it is shown how to keep $n^* \gg 1$ in extreme cases, by repeating certain lines of code with a reduced $\alpha$.
In addition, VA-Flow can be modified with the heuristic of including momentum on the updates of ${\bm \theta}$.
(Although, it will be shown that momentum isn't needed as much as it is with basic GD, since the updates are more accurate.)
Also, there are a number of coding improvements that should be utilized in a real implementation, but which
weren't shown in Algo.~\ref{algo:va-flow} for simplicity.
Examples include checks to avoid dividing by zero; see the implementation by \cite{MFZ-github}.

\section{Simple Demonstration}

\begin{figure}[ht!]
\begin{center}
\includegraphics[width=0.5\linewidth]{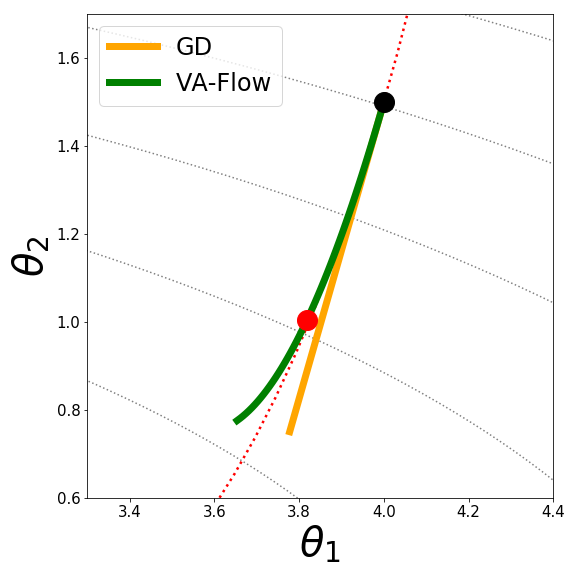}
\captionof{figure}{ 
A close-up view of the updates for basic GD (\underline{orange}) and VA-Flow (\underline{green}) for a range of $\alpha$ values when seeking the minimum of an ellipse function.
The light \underline{gray} lines are contours of the CF, and the light \underline{red} line is the exact path toward the minimum.
The \underline{black dot} is the starting point for both, and the \underline{red dot} is the recommended update from VA-Flow.
}
\label{fig:ellipse}
\end{center}
\end{figure}

The purpose of this section is to illustrate the difference between Basic GD and the new VA-Flow algorithm
on a simple example, that of the optimization of the ellipse cost function $f$
\begin{align*}
f & = \frac{\theta_1^2}{c_1^2} + \frac{\theta_2^2}{c_2^2} \, \, \, .
\end{align*}
This comparison will be illustrated using a {\em single update} of ${\bm \theta}$.
Following Eq.~\ref{eqn:update2}, the update for ${\bm \theta}$ 
for Basic GD (i.e., ${\bm \theta}_{GD}$), and for VA-Flow (i.e., ${\bm \theta}_{Flow}$) appear as the following functions of $\alpha$
\begin{align*}
{\bm \theta}_{GD} & = {\bm \theta}_0 + \alpha {\bm v} \\
{\bm \theta}_{Flow} & = {\bm \theta}_0 + \alpha {\bm v} + \frac{\alpha^2}{2} {\bm a} \, \, \, ,
\end{align*}
where ${\bm \theta}_0$ is the initial condition for the parameter.
Note that ${\bm \theta}_{Flow}$ has an additional quadratic $\alpha$-dependence that ${\bm \theta}_{GD}$ does not have; this allows it to more closely follow the exact path from infinitesimal updates.
Using the parameter choices $c_1 = 6$, $c_2 = 2$ and ${\bm \theta}_0^T = (4, 1.5)$,
it follows
\begin{align*}
{\bm v}^T & = -2 (\theta_1 / c_1^2, \theta_2 / c_2^2 ) = (-0.22..., -0.75) \\
{\bm a}^T & = 4 (\theta_1 / c_1^4, \theta_2 / c_2^4 ) = ( -0.012..., 0.375 ) \, \, \, ,
\end{align*}
from which both ${\bm \theta}_{GD}$ and ${\bm \theta}_{Flow}$ are plotted as functions of $\alpha$ in Fig.~\ref{fig:ellipse}.
The green line represents ${\bm \theta}_{Flow}$, while the orange line is ${\bm \theta}_{GD}$.
Also, ${\bm \theta}_0$ is depicted with a black dot.
Using Approach A and requiring $\rho_{targ} = 0.2$, it follows $\alpha_A \approx 0.834$.
Using $\alpha_A$, the recommended ${\bm \theta}_{Flow}$ for a single update can be computed (following Approach A)
and appears as a red dot in the figure.
The dotted red line is the exact path from infinitesimal GD updates; it is described by $\theta_2 \sim \theta_1^c$ with $c = (c_1/c_2)^2$, 
as shown in the  Supplement.
This figure makes clear the advantages VA-Flow has to offer: a means to determine a best value of $\alpha$,
and a more accurate update of ${\bm \theta}$.

\section{Application: Inverse Kinematics}

\begin{figure}[ht!]
\begin{center}
\includegraphics[width=0.8\linewidth]{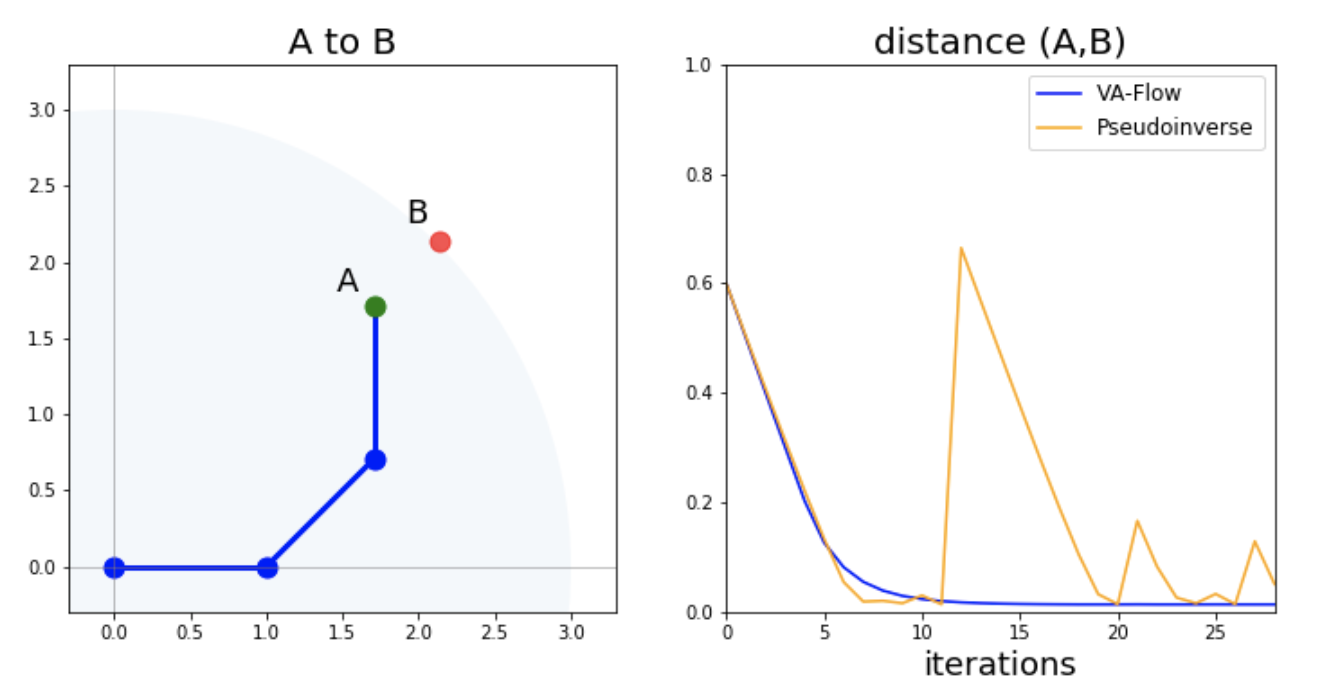}
\captionof{figure}{ 
In the left plot, point B (i.e., the \underline{red dot}) is the target, while point A (i.e., the \underline{green dot}) is the starting location for the end effector.
The three \underline{blue dots} are the joints of the arm, and are parametrized by ${\bm \theta}$.
The light blue shading indicates the range of motion of the arm.
The right plot graphs the distance between ${\bm s}_A$ and ${\bm s}_B$ for each algorithm, revealing a smooth and efficient approach for {\em VA-Flow} and an erratic approach for the {\em Jacobian Pseudoinverse} method.
}
\label{fig:IKarm}
\end{center}
\end{figure}

This application is based on a 3-link arm, where the links (each of length $1.0$) are constrained to move in a 2D plane.
The parameter vector ${\bm \theta} = (\theta_1, \theta_2, \theta_3)$ describes the angles for the three joints, indicated
by blue dots in the left plot of Fig.~\ref{fig:IKarm}; there are no constraints on these angles.
A simple test suffices to illustrate the difference between an arm controlled by {\em VA-Flow} (using Approach A) and the method of {\em Jacobian Pseudoinverse} discussed earlier.  
The test is merely to move the end-effector from point A to B, starting from the initial
pose shown in the left plot of Fig.~\ref{fig:IKarm}; the two points are ${\bm s}_A = (1.71,1.71)$ and ${\bm s}_B = (2.132,2.132)$.
Point B is just beyond the range of motion of the arm.  
The results for each algorithm are shown in the right plot of the figure, which measures the distance 
$\| {\bm s}_B - {\bm s}_A \|$ versus iterations.  
It reveals that VA-Flow leads to a very smooth approach to the target, with no oscillations or erratic behavior. 
In contrast, the graph for the basic Jacobian Pseudoinverse approach reveals very erratic updates: as the end effector becomes close to the target, it suddenly moves away.  
In general, it was observed that the Jacobian Pseudoinverse algorithm performed well when the target was within the range of motion, but poorly outside this range.
Also, it performs poorly when the arm is near a singular configuration (i.e., when two or more rows of the Jacobian are nearly equal).
In contrast, the VA-Flow algorithm performed well in all circumstances.
The instability was not manifested with VA-Flow because it is a second-order algorithm, and can better navigate such singular configurations.
Also, as pointed out earlier, second-order methods based on the quasi-Newton method do not exhibit erratic behavior either.
Although those methods are computationally much more costly in comparison to VA-Flow. 
 
Finally, there were some minor adjustments made to the VA-Flow algorithm for the IK application; these are discussed in the Supplement.
Primary among them is that $\alpha$ should be limited by (1) the maximum desired update size of the end-effector, and (2) the remaining
distance to the target.

\section{Application: Gradient Descent}

\begin{figure}[ht!]
\begin{center}
\includegraphics[width=0.9\linewidth]{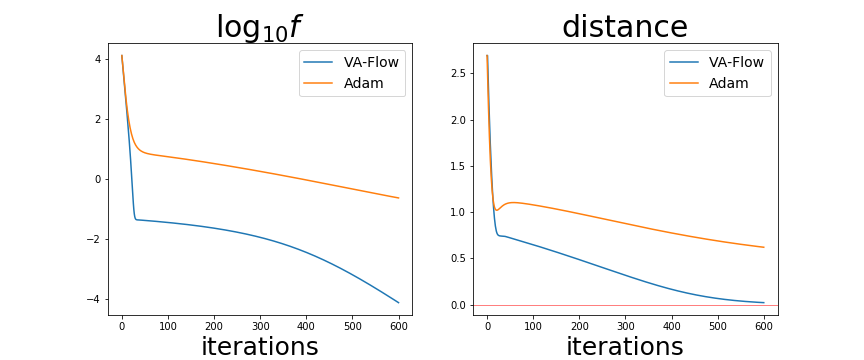}
\captionof{figure}{ 
A comparison figure for the {\em VA-Flow} and {\em Adam} algorithms.
In the left plot, the log of the CF is plotted for each algorithm, revealing that VA-Flow reaches a value $10^{-4}$ the size of the CF reached by {\em Adam}.
The right plot measures the distance from the current ${\bm \theta}$ to the known global minimum, and likewise reveals superior performance by VA-Flow.
}
\label{fig:GD}
\end{center}
\end{figure}

The VA-Flow algorithm can largely be applied to a GD example in the same way as for IK.
Here the example is to determine the minimum of Beale's function, defined by 
\begin{equation*}
f = (1.5 - \theta_1 + \theta_1 \theta_2 )^2 + (2.25 -\theta_1 + \theta_1 \theta_2^2)^2 + (2.625 - \theta_1 + \theta_1 \theta_2^3)^2 \, \, \, .
\end{equation*}
The known global minimum of this function is $(\theta_1,\theta_2)=(3,0.5)$;
the initial point used is $(4,3)$, from which the minimum is reachable.
In Fig.~\ref{fig:GD} a comparison is made between VA-Flow and Adam on this CF.
For VA-Flow, the learning rate was $\alpha = 4.8 \times 10^{-6}$, Approach A was used, and $M=100$.
Also, no momentum was used for VA-Flow.
For Adam, $\alpha$ was set to $0.15$, and the other parameters had default settings.
The most obvious difference between these two algorithms is that, as shown in the left plot, VA-Flow is able to reach much lower CF values, which is the desired result.  Also, the right plot displays the distance between the current ${\bm \theta}$ and the target, also revealing the superior performance of VA-Flow.

Another example investigated by the author was using VA-Flow to minimize the cross entropy CF,
which was defined on a neural net (NN) model for classifying handwritten digits \citep{scikit-learn,scikit-digits,Dua2019}.
The NN had one hidden layer with 30 nodes; there were 64 input nodes and 10 output nodes.
The CF on the final layer was a cross entropy cost function using {\em softmax}, and the hidden layer used a {\em tanh} activation function.
Altogether, the dimensionality of ${\bm \theta}$ was 2260.
As it turned out, the VA-Flow algorithm fared poorly in the initial iterations, since $\| {\bm a} \| \ll \| {\bm v} \|$.
This led to a large $n^*$ (and $\alpha$) and subsequently poor behavior.
In retrospect this is perhaps not surprising.
Even though the VA-Flow algorithm is using 2nd-order information, it is purely {\em local} information.
Such localized information is of limited value on the cross entropy CF, which is composed of exponentials,
and is characterized by very flat and very steep regions.
A better approach is one that contains information over a {\em neighborhood} of values surrounding the current ${\bm \theta}$, such as the Neograd algorithm \citep{MFZ-neograd}.

\section{Discussion}

A method has been introduced to efficiently calculate the rate of change of a vector field, and in so doing,
compute with ${\cal O}(N)$ complexity a task which has long been construed to be an ${\cal O}(N^2)$ task.
From this, the algorithm VA-Flow was created, which was based on the philosophy that the correction to a first-order update
should be smaller than the first-order update itself.
The value of the parameter $\alpha$ determined by this algorithm can be used within a first or second order version of the algorithm.
It is noted that a numerical estimate of the quantity ${\bm a}$ was previously done by \cite{Moller1993,Moller1993-thesis} and \cite{Pearlmutter1994}.
However, they made different applications of it, and in particular did not derive the 
second order update formula for ${\bm \theta}$ (Eq.~\ref{eqn:update2}), which was essential for the VA-Flow algorithm.

When applied to inverse kinematics (IK), the algorithm proved to be extremely successful.
It was capable of moving an articulated limb in ${\cal O}(N)$ time, without any erratic movement whatsoever.
Before VA-Flow, that was only capable with algorithms that required ${\cal O}(N^2)$ computations.

VA-Flow was also applied to a polynomial test function, and performed much better than Adam,
reaching a CF that was about $10^4$ times smaller than that reached with Adam.
However, when it was applied to a more realistic application (a modestly sized NN), it didn't fare as well.  
The author reasons it is because the VA-Flow places too much importance on measurements made approximately at the same point.
While this works well on polynomials, it is perhaps not well-suited for functions involving steep and flat regions caused by exponentials.
An algorithm such as Neograd \citep{MFZ-neograd} seems to be a better choice.

There are a number of ways to move forward with the results presented here.
First, one might consider an additional measurement of ${\bm v}({\bm \theta})$ in the neighborhood of ${\bm \theta}_0$, in order to assess the rate of change of ${\bm a}$ itself.
Another approach is to assume that ${\bm v}$ changes in an exponential manner, and model it appropriately (perhaps using two or three
measurements of ${\bm v}$).  This could be helpful when working with CFs from NNs.

Finally, the mathematical results presented in this paper have myriad applications with varying ethical impacts.
Ethical considerations were made on the IK and GD applications in Sec.~\ref{sec:background}.









\bibliography{MZproj15}

\end{document}